# Stochastic-based pattern recognition analysis

Vincent Canals, Antoni Morro and Josep L. Rosselló*,

Electronic Systems Group. Physics Department. Universitat de les Illes Balears

Edifici Mateu Orfila. Cra. Valldemossa km. 7.5. 07122 Palma de Mallorca, Balears, Spain

* Corresponding author: Tel. +34 971 171 373. Fax. +34 971 173 426. email:

j.rossello@uib.es

**Abstract: In this work we review the basic principles of stochastic logic and propose its application to probabilistic-based pattern-recognition analysis. The proposed technique is intrinsically a parallel comparison of input data to various pre-stored categories using Bayesian techniques. We design smart pulse-based stochastic-logic blocks to provide an efficient pattern recognition analysis. The proposed architecture is applied to a specific navigation problem.**

# 1. Introduction

Stochastic logic is the result of applying probabilistic laws to logic cells (Gaines, 1968) where variables are represented by random pulse streams. Stochastic computing makes use of digital technology to perform arithmetic operations with the advantage of requiring a reduced number of gates. Stochastic logic uses pulsed signals that represent different quantities depending on its switching probability. Pulses can be converted to binary numbers by using digital counters (P2B converters) while binary numbers can be translated to stochastic signals by combining a random (or a pseudo-random) number generator and a comparator (B2P converters) (see the work of Kim and Shanblatt, 1995). Recently stochastic pulse sequences have been used to implement a deterministic logic (Bezrukov and Kish 2009).

Stochastic computing is useful for those applications requiring parallel-processing techniques. Traditional parallel processing architectures have the shortcoming of requiring a large amount of hardware and computation tasks are relatively extensive in complexity. Therefore, the number of tasks that are executed in parallel is reduced in number. Stochastic computing could represent a solution to this problem since the hardware associated to each task is reduced in size if compared to traditional digital implementations. Therefore, more complex tasks can be executed in parallel when using stochastic computing elements.

In this work we present a low cost methodology for pattern-recognition. For this purpose we selected stochastic logic due to the low number of gates needed with respect to the complexity of operations being involved (such as arithmetic multiplication, addition, division, ...). The probabilistic nature of stochastic logic is ideal to perform pattern recognition using bayesian techniques (Bishop, 2006). The proposed approach is verified showing a simple example and also applying the method to the navigation of an autonomous vehicle. The autonomous vehicle is able to measure the distance to nearest objects and, using stochastic computing, recognize the environment.

This paper is organized as follows: in Section 2 we briefly explain the stochastic computing concept, in Section 3 we show a methodology to apply stochastic logic for pattern recognition. In Section 4 we apply the proposed solution to the motion of an autonomous vehicle, and finally we expose the conclusions in Section 5.

## 2. Basic Principles of Stochastic Logic

*2.1 Introduction to Stochastic Logic*

In stochastic-based computations a global clock provides the time interval during which all stochastic signals are stable (settled to 0 or 1). During a clock cycle, each node has a probability *p* of being in the HIGH state (see Fig. 1). This probabilistic-based coding provides a natural way of operating with analogue quantities (since probabilities are defined between 0 and 1) using digital circuitry.

Pulsed signals follow probabilistic laws when they are evaluated through logic gates. As an example, the AND gate provides at the output the product of their inputs (that is, the collision probability between signals), and a NOT gate converts the probability *p* at the input to the complementary *1-p* at the output. The sum of two switching signals may be implemented using a multiplexer while the division of two numbers can be obtained by using an up/down counter and a binary to pulse converter (B2P) to iteratively estimate the pulsed signal h=p/q such that h•q=p (see Fig. 2).

One of the requirements of stochastic computing is that signals must be un-correlated at different clock cycles. In Fig. 3 we show the importance of the temporal un-correlation when cascading arithmetic functions (we use the example of implementing *f(p)=p(1-p)*). In the first circuit the output is always equal to zero since signals *p* and *1-p* at the inputs of the AND gate are correlated and therefore the output has always a LOW value. Such a correlation can be eliminated using shift registers to delay signals from one arithmetic level to the next as is shown in Fig. 3. In the correct case, the AND gate always evaluates the product between *p* and a delayed (and therefore uncorrelated) value of *1-p*.

*2.2 Evaluating time and error estimation in signal translation*

Pulsed signals can be converted to binary numbers using counters. We define a pulse to binary converter of order *N* (a P2B(N)) as those circuits that evaluate the number of HIGH values provided by a stochastic signal during *N* clock cycles. The output of a P2B(N) block is an n-bit number that changes every *N* cycles so that the evaluation time is $T_{EVAL}=N \cdot T$ (where *T* is the clock cycle).

An error is always present during conversions that can be probabilistically estimated. We define $P_N(x)$ as the probability of the P2B(N) output to be equal to the binary number *x*. This probability follows a binomial distribution:

$$P_N(x) = \binom{N}{x} p^x (1-p)^{N-x} \tag{1}$$

where parameter '*p*' is the switching probability of the input signal. The mean value provided by the converter is $<x>=Np$ (the desired result) while the standard deviation of the distribution is $\sigma^2=Np(1-p)$. Defining the relative error during each conversion as $Error \equiv 2\sigma_{max}/N$, and considering that the maximum deviation ($\sigma_{max}$) happens when $p=1/2$, then $Error \sim 1/N^{1/2}$. The relationship between error and conversion time ($T_{EVAL}$) is therefore given by the next expression:

$$T_{EVAL} = NT = T\left(\frac{1}{Error}\right)^2 \tag{2}$$

Equation (2) implies a trade-off between conversion time and error. Small errors imply large conversion times and inversely fast conversions imply large errors.

## 3. Stochastic-Based Pattern Recognition

The probabilistic nature of stochastic logic is an advantage for the implementation of probabilistic-based pattern recognition methodologies. The purpose is to compare signals coming from different sensors (the features) with reference values that represent different categories. Two parameters (the mean and the sigma value) are required to define each category zone assuming a normal probability density function (that is, the likelihood function of the category with respect to the input signal). Fig. 4 shows a stochastic circuit used to generate a normal probability density function with mean value $\eta$ and a dispersion $\sigma_{ref}$ dependent on the integration time N. A pulsed signal, (the feature signal) is evaluated through a P2B(N) converter. Since the output of such digital block follows the binomial distribution (1) the probability of coincidence between the output of the block and the mean value also follow a binomial distribution (equivalent to a normal distribution if N is sufficiently large). Therefore, the switching activity at the output of the comparator in Fig.4 is proportional to the probability of the feature to be within the category defined by a binomial distribution centred at the reference signal ($\eta$) (see the two cases shown in Fig. 4). From basic theory of probabilities, the relative dispersion of the distribution ($\sigma_{ref}/N$, similar to the relative error computed in the previous section) is found to be inversely proportional to the square root of the value selected for N ($\sigma_{ref}/N \approx p(1-p)/\sqrt{N}$). Therefore high (low) values of N imply a low (high) relative dispersion.

If we are interested in the generation of the *a posteriori* probability function we must implement the Bayes formula:

$$P(C_i|x) = \frac{P(C_i)P(x|C_i)}{\sum_j P(C_j)P(x|C_j)} \quad (3)$$

where P(C$_i$|x) is the *a posteriori* probability, that is the probability of being of class Ci given we measure x. P(C$_i$) are the *a priori* probability of class C$_i$ and P(x|C$_i$) are the known likelihood functions of C$_i$ with respect to x (probability of measuring 'x' given is of the class C$_i$).

A simple example of two-class problem has been applied using the stochastic circuit shown in Fig. 5. Starting with the measurement 'x' (a pulsed signal) we generate the likelihood PDF functions P(x|A) and P(x|B) that are operated with the *a priori* probability of classes P(A) and P(B)=1-P(A) to estimate the *a posteriori* probability function P(A|x) using the Bayes formula:

$$P(A|x) = \frac{P(A)P(x|A)}{P(A)P(x|A) + P(B)P(x|B)} \quad (4)$$

Results of the circuit implemented are shown in Fig. 6 where we compare circuit measurements (symbols) and the expected behaviour (lines) for $\eta_A$=0.5, $\eta_B$=0.3, $\sigma_A$=$\sigma_B$=0.08 and P(A)=0.5.

The extrapolation to more than one feature (implementation of the joint probability distribution function) can be done using different normal PDF generators working in parallel (one for each sensor signal that are assumed to be independent). We define those blocks as "*Multi-dimensional Stochastic Classifiers*" (MSCs). In Fig. 5 we show an example of those MSC blocks. Note that an MSC bloc would provide a stochastic signal with a switching activity proportional to the suitability of the n-dimensional feature vector of being included in a given category (that is, the likelihood function of C$_i$ with respect to x

P($\mathbf{x}$|$C_i$)). We use as many MSC blocks in parallel as categories are defined in the system. We assumed the case in which the *a priori* probabilities of classes are equal and therefore the recognition of classes are only dependent on the likelihood PDF functions (thus further simplifying the circuit). The MSC blocks are constructed following a competitive scheme so that when the internal counter of an MSC block arrive to the end ($x_i$=1) then the internal counters of the other MSC blocks (of the other categories) are reset.

We can compare the performance of the proposed methodology (implementing it in an FPGA) with respect to software solutions (use of a microprocessor) and the use of classical digital electronics (using an FPGA). In Table 1 we show the comparison between the three methodologies. We implemented the comparison between two different classes using 1-dimensional MSC blocks (with only one feature). Due to the low cost in terms of gates of the proposed stochastic solution we can implement since 650 different 1-dimensional comparators in an ALTERA STRATIX II EP2S60 IC. Using traditional digital electronics we must implement the comparison using different multipliers, dividers, adders and comparators along with an embedded memory (to implement a look-up table for the exponential function in order to implement the Gaussian likelihood PDF functions). In the selected IC we can only implement one single digital comparator in the whole IC. Although being faster than the stochastic solution (about 70 times faster), the total performance of the proposed solution is one order of magnitude faster than classical implementation due to the advantages of parallelism. A higher performance improvement is observed (a factor of 3000) when compared with software solutions, where computations must be performed sequentially.

## 4. Results

We verified the proposed patter-recognition technique to orient an autonomous vehicle in a known environment. At each time step, the vehicle computes the distance to the nearest walls (parameters n,s,e and w, in Fig. 8). These parameters are the inputs of seven MSCs (one for each zone of the plane) configured with specifics mean and sigma values (parameters $\eta$ and N in Fig. 4). Once the vehicle recognizes the environment (the class recognized is the zone where it is the vehicle), a predefined movement is selected with the objective to reach zone A (see Fig. 8). In different experiments were the vehicle was placed randomly in the plane, the time steps needed to reach zone A were computed. In all cases, the vehicle was able to reach zone A with a nearly minimum trajectory. In Fig. 9 we compare the time steps required to reach zone A (symbols) with the optimum number of time steps (solid line). Results show that the itinerary followed by the vehicle is very close to the optimum trajectory thus demonstrating that the stochastic system is able to recognize the zones at each point of the path followed.

## 5. Conclusions

In this paper we reviewed the main characteristics of stochastic logic. The associated error for stochastic-to-binary conversions has been estimated and a methodology to implement stochastic computations in cascade has been presented. A complete scheme of a stochastic divider is presented for the first time and also a new methodology of probabilistic-based pattern recognition analysis using stochastic computing. The proposed methodology may represent a low-cost solution for an efficient hardware implementation of pattern classification since massive parallel-connected stochastic structures can be built on a single chip. The new pattern recognition methodology has been applied to the navigation of an autonomous mobile vehicle.

**References**

Bezrudov, S.M., and Kish, L.B., 2009. Deterministic multivalued logic scheme for information processing and routing in the brain. Physics Letters A, **373** (27-29), 2338-2342

Bishop, C.M., 2006. Pattern recognition and machine learning. Springer, New York

Gaines, B.R., 1968. Random pulse machines. IEEE Trans. on Comp. 410-410

Kim, Y.K., and Shanblatt, M., 1995. Architecture and statistical model of a pulse-mode digital multilayer neural network. IEEE Trans. on Neural Networks, **6** (5), 1109-1118

**Figure captions:**

Fig. 1. Basic concepts for stochastic signals

Fig.2. Simple logic gates can be used to perform arithmetic computations when using stochastic signals

Fig. 3. Signal correlation impact on stochastic arithmetic blocks

Fig. 4. General design of a stochastic normal PDF generator

Fig. 5. Circuit used to estimate the *a posteriori* PDF function

Fig. 6. Results of the simple stochastic Bayesian estimator

Fig. 7. Multi-dimensional stochastic comparator (MSC). Output signal "Category (i)" is equal to '1' if sensor signals are within the i-th category.

Fig. 8. Plane and movement map of the autonomous vehicle

Fig. 9. Time steps needed by the vehicle to reach zone 'A' from different points of the plane.

────    Minimum path

◊    Paths followed

**Table captions:**

Table 1. Comparison between stochastic, classical digital electronics and software implementations. The proposed method is three orders of magnitude faster than software and one order of magnitude faster than classical implementation.

**Figure 1:**

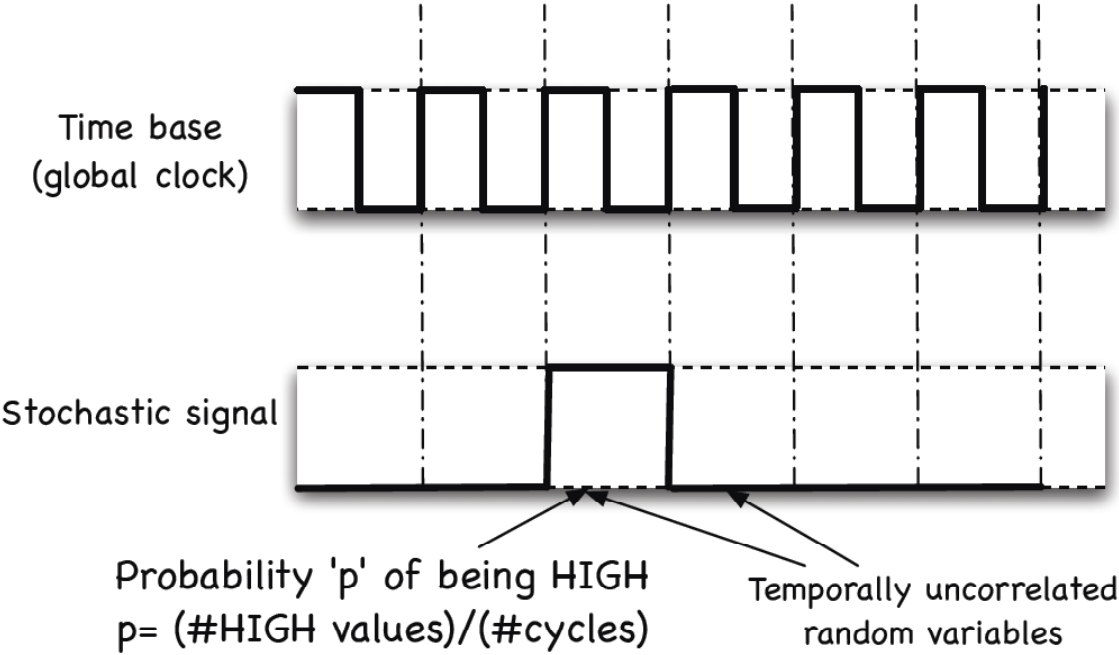

**Figure 2:**

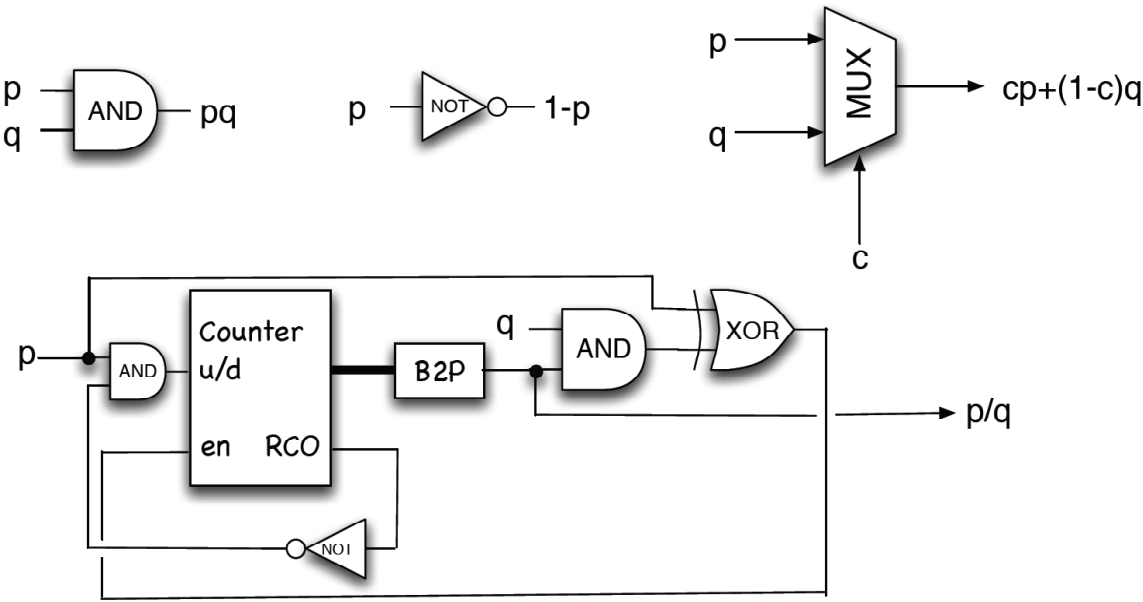

**Figure 3**

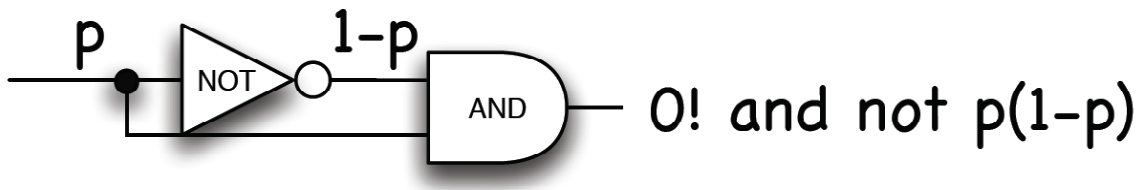

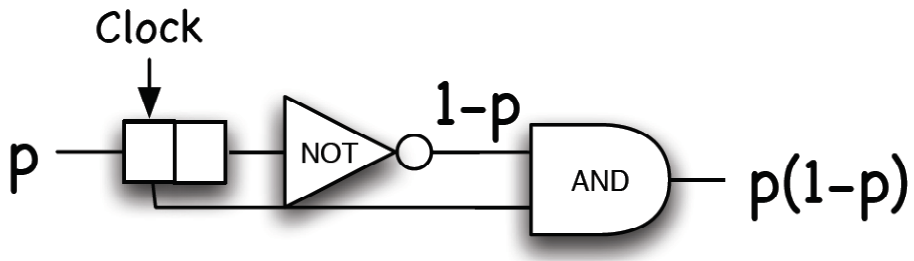

**Figure 4:**

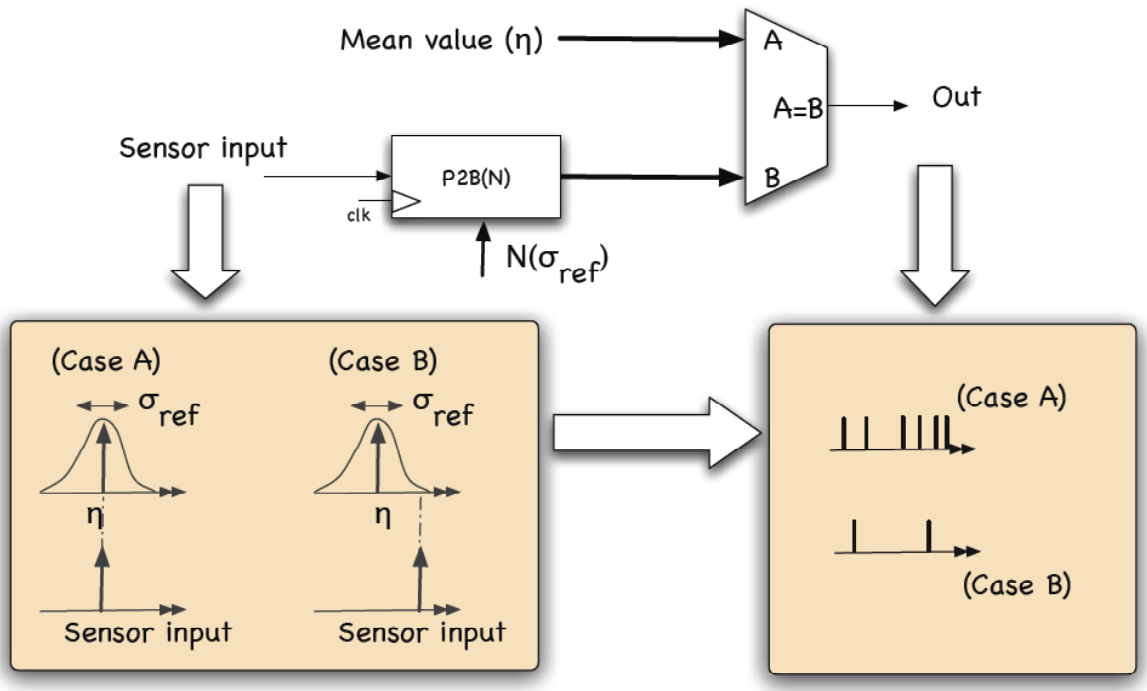

**Figure 5:**

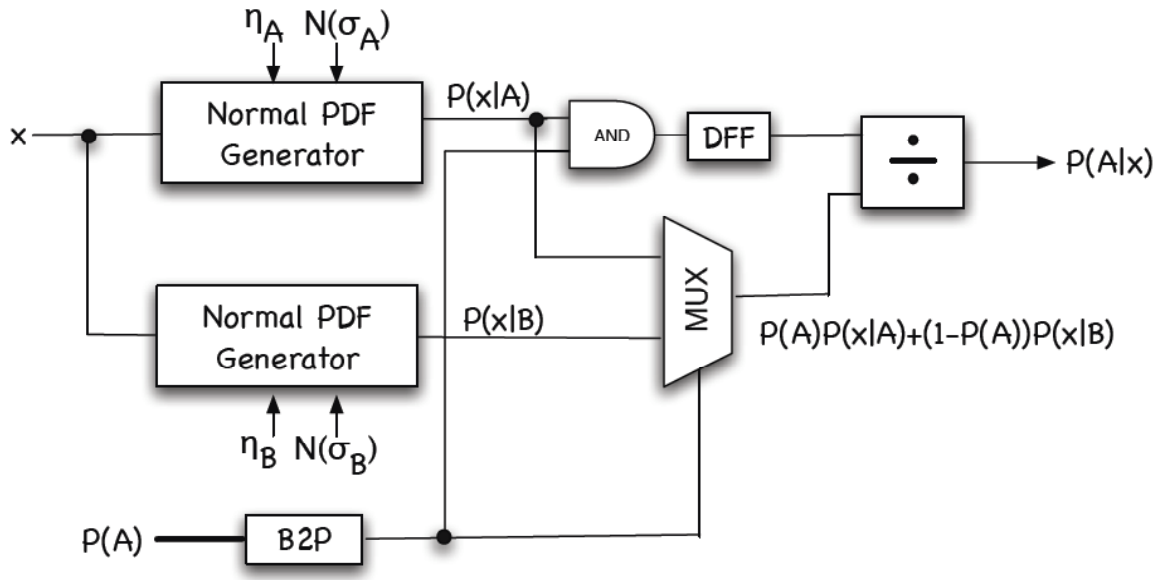

**Figure 6:**

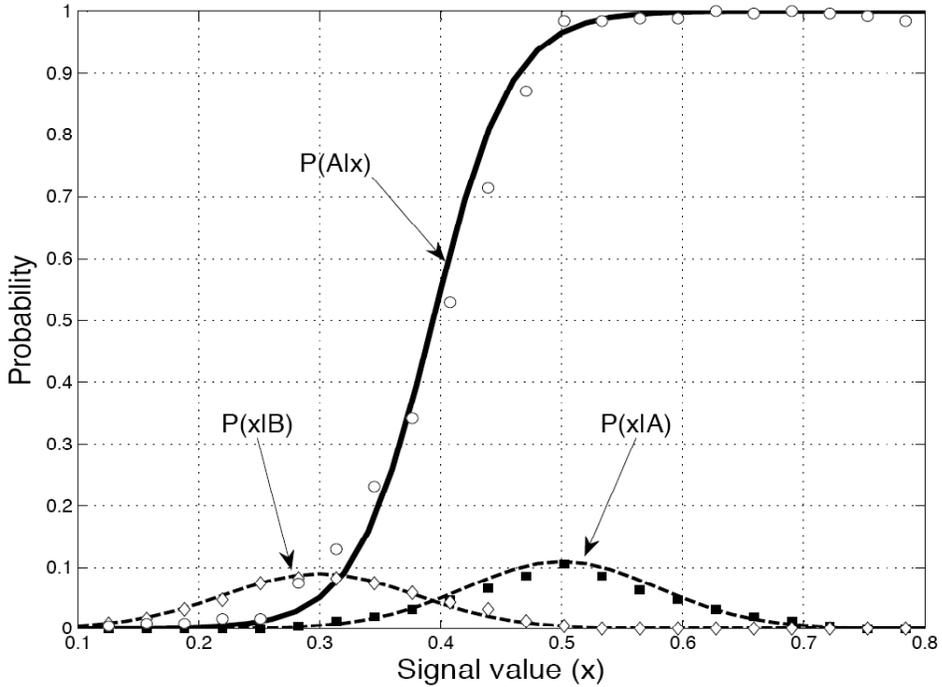

**Figure 7:**

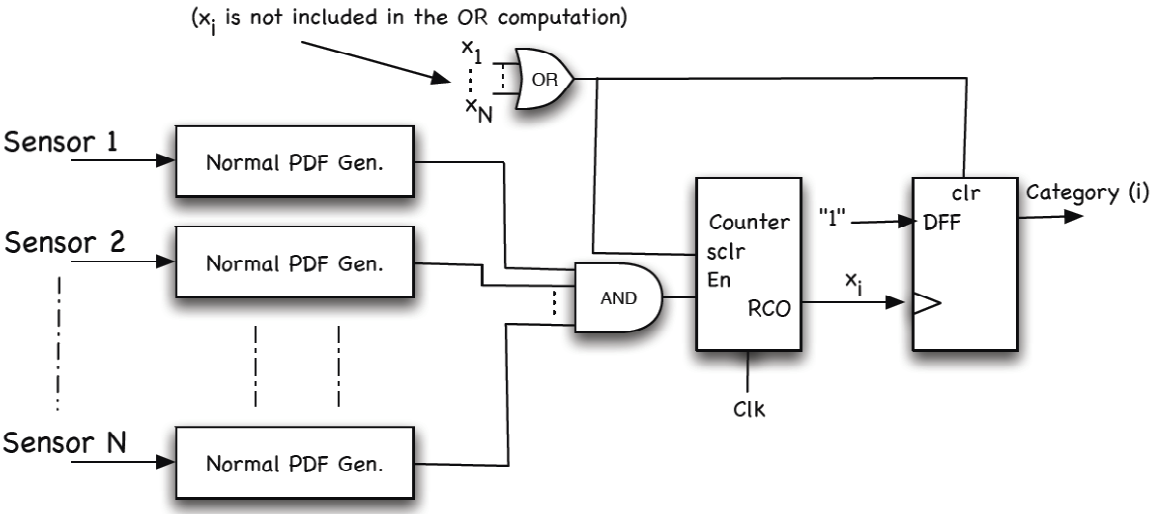

**Figure 8:**

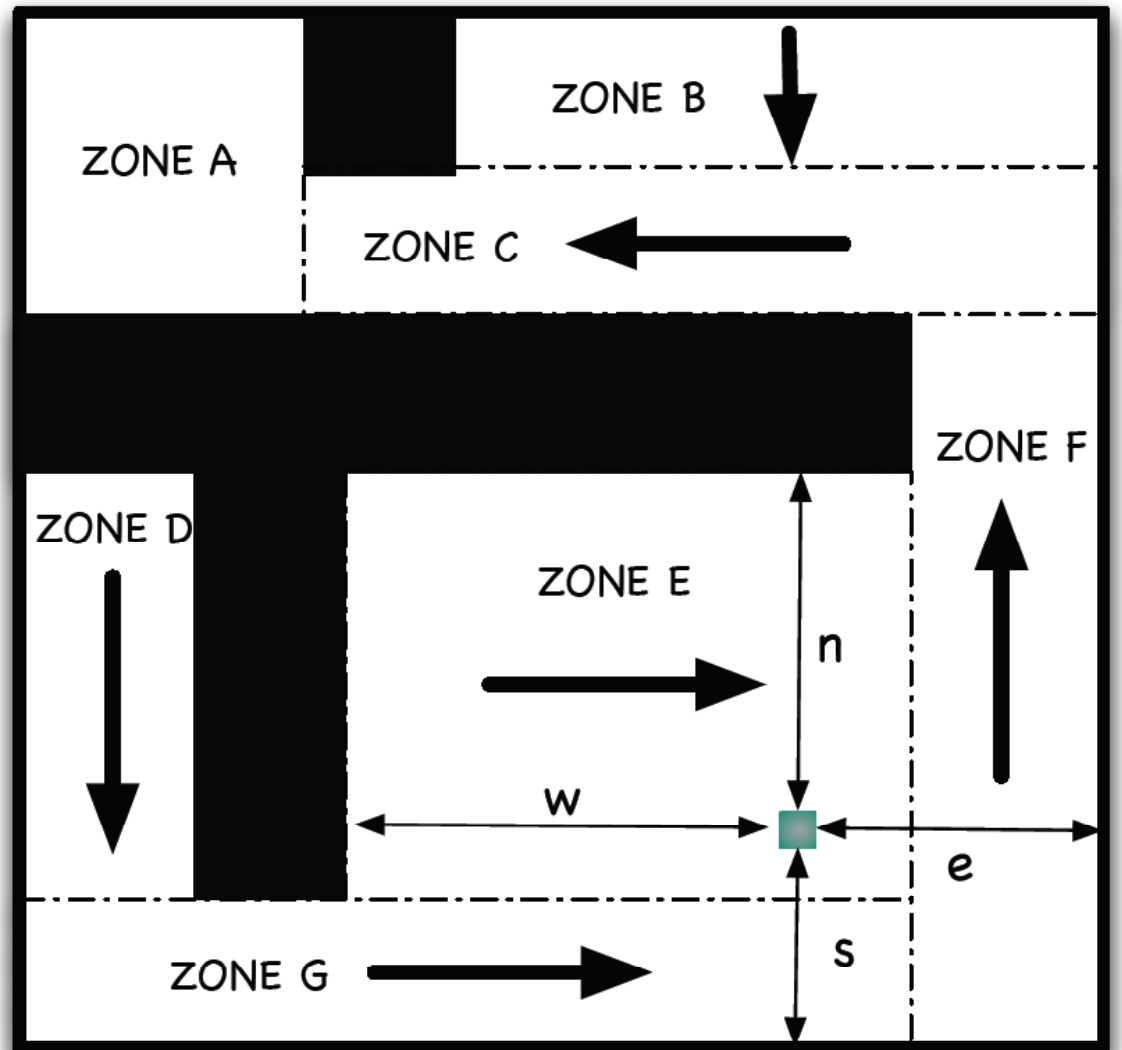

**Figure 9:**

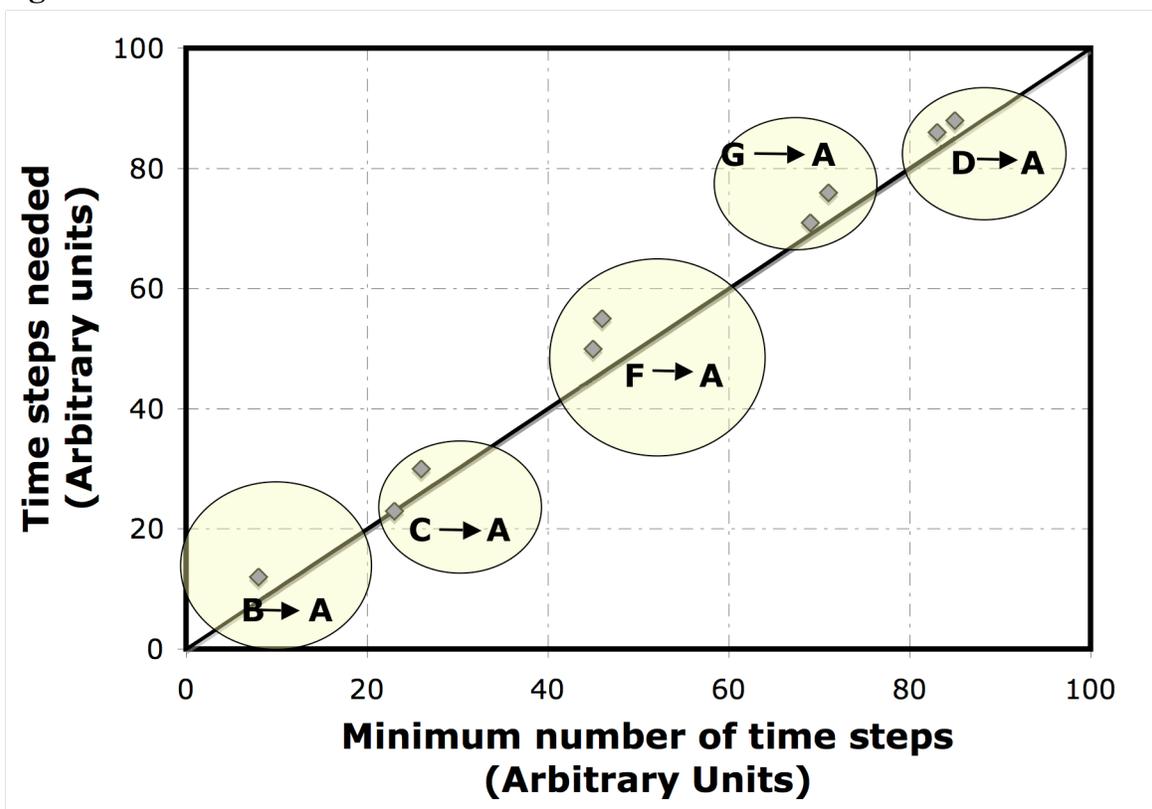

**Table 1:**

|  | Stochastic methods using FPGAs | Classical digital electronics using FPGAs | Microprocessor |
|---|---|---|---|
| **IC** | Stratix II EP2s60 | Stratix II EP2s60 | Intel Q9400 |
| **Technology** | 90nm | 90nm | 45nm |
| **Operating Frequency (GHz)** | 0,45GHz | 0,45GHz | 2,66GHz |
| **Cost** | 1325,00 USD | 1325,00 USD | 260,00 USD |
| **Numumber of Processing Elements per chip** | 650 | 1 | 1μP |
| **Propagation delay per computation (μs)** | 7 | 0,113 | 30,55 |
| **Performance (Millions of Comparisons per secon)** | 92,86 | 8,85 | 0,03 |